\documentclass[iicol,pdflatex,sn-basic]{sn-jnl}

\usepackage{graphicx}%
\usepackage{multirow}%
\usepackage{amsmath,amssymb,amsfonts}%
\usepackage{amsthm}%
\usepackage{mathrsfs}%
\usepackage[title]{appendix}%
\usepackage{xcolor}%
\usepackage{textcomp}%
\usepackage{manyfoot}%
\usepackage{booktabs}%
\usepackage{algorithm}%
\usepackage{algorithmicx}%
\usepackage{algpseudocode}%
\usepackage{listings}%
\usepackage{adjustbox} 

\theoremstyle{thmstyleone}%
\theoremstyle{thmstyletwo}%

\theoremstyle{thmstylethree}%

\begin{document}

\title[Article Title]{Improving Robustness of Foundation Models in Domain Adaptation with Soup-Adapters}


\author*[1]{\fnm{Marco} \sur{Roschkowski}}\email{roschkowski@uni-wuppertal.de}

\affil*[1]{\orgdiv{School of Mathematics and Natural Sciences}, \orgname{University of Wuppertal}, \orgaddress{\street{Gaußstraße 20}, \postcode{42119}~\city{Wuppertal}, \country{Germany}}}


\abstract{In this paper, we tackle two fundamental problems in few-shot domain adaptation
of foundation models.
First, hyperparameter tuning is often impractical due to the lack of large validation datasets. Second, model robustness under distribution shifts where test time data deviates slightly from training distributions, remains a concern.
We show that by training multiple independent adapters and averaging their outputs, the new model has a higher performance and is more robust to distribution shifts compared to any individual adapter.
This improvement holds even when the adapters are trained with diverse hyperparameters sampled from a wide range, resulting in varied individual performance.
Consequently, our method addresses both of the problems described above.
The ensemble is also significantly less sensitive to the residual ratio, a critical hyperparameter of CLIP-Adapter.
Since the ensemble can be reparameterized to a single adapter again using a principled concatenation of the parameters, we refer to our method as Soup-Adapter. 
This is also the first study to explore CLIP adapter-style techniques for DINOv2 and to directly compare them with CLIP in this setting.
}




\maketitle
\section{Introduction}
Computer vision has seen tremendous progress due to the emergence of deep learning technologies. Large supervised benchmark datasets such as ImageNet \citep{deng2009imagenet} have led to several methodological breakthroughs. These include overcoming traditional computer vision methods in \citep{krizhevsky2012imagenet}, the introduction of skip connections in \citep{he2016deep}, advanced architectures such as inverted bottlenecks in \citep{sandler2018mobilenetv2} and improved scaling techniques in \citep{koonce2021efficientnet}.

A~long-standing limitation has been the dependence on such large curated datasets which are expensive to obtain. Recently, the paradigm of foundation models has become an attractive alternative in which a single model is being trained on a corpus of data large enough to generalize well on several distinct downstream tasks. One notable vision foundation model is CLIP \citep{radford2021learning} which learns a joint embedding space of images and their corresponding captions.
This architecture naturally has the ability to perform zero-shot classification by describing visual categories via text prompts.
Another popular foundation model is DINOv2 \citep{oquab2023dinov2} which has been trained on a large curated corpus of images to produce robust features.
These models can be easily adapted for few-shot learning using KNN evaluation or prototypical learning \citep{snell2017prototypical}.
However, CLIP models do not achieve the same zero-shot performance as state-of-the-art supervised models.
Similarly, DINOv2 models do not achieve the same KNN-evaluation accuracy as comparable fully supervised models.

This has led to a growing interest in few-shot adaptation techniques where the goal is to boost the performance of foundation models on the specific downstream task using a small number of additional labeled samples. In CoOp \citep{zhou2022learning}, it has been shown that the learning of optimal text prompts can improve downstream performance.
Adapters such as the CLIP-Adapter \citep{gao2024clip} have become a dominant line of research where the foundation model itself is kept frozen and only additional parameters are being learned.
CLIP adapters typically have a very low number of parameters compared to the foundation model itself.

This work aims to address two challenges that are common in few-shot learning with adapters. The first problem is that it is often not possible to perform extensive hyperparameter tuning because large validation datasets are typically not available. A particularly important hyperparameter, called the residual ratio, is often necessary if CLIP-Adapters overfit. After training the adapter, this balancing factor allows one to smoothly interpolate between the original zero-shot features and the trained adapter features. 
Secondly, our goal is to improve performance under distribution shifts of adapter networks.
Our contributions may be summarized as follows. \begin{itemize} \item We propose Soup-Adapter, a simple yet effective ensemble of independently trained CLIP-Adapters, which averages their predictions to improve performance. \item We demonstrate that Soup-Adapter not only enhances the accuracy of CLIP-Adapter but also significantly improves robustness under distribution shifts, an important limitation of many domain adaptation methods.
\item Since Soup-Adapter works well when the individual hyperparameters are sampled randomly from a diverse range of configurations for each component, we effectively tackle the problem of hyperparameter tuning.
\item The proposed method is less sensitive to the residual ratio.
\item We demonstrate that CLIP-Adapter-based methods, especially the proposed Soup-Adapter, perform effectively with DINOv2 models.
\item We show that Soup-Adapter can be reparameterized into a single adapter—sharing the same architecture as the individual adapters in the ensemble—through parameter concatenation.
\end{itemize}
The structure of this work is as follows.
We will discuss related work in Section~\ref{sec:related_work}. Our method will be introduced in Section~\ref{sec:method}. We thoroughly evaluate our method on several datasets in Section~\ref{sec:experiments} and conclude the paper with Section~\ref{sec:conclusion}.

\section{Related Work}\label{sec:related_work}
\subsection{Emergent Properties of Foundation Models}
GPT-3 \citep{brown2020language} demonstrated that large language models trained on massive corpora have impressive few-shot capacities and generalize well on very different language tasks. 
In DINO \citep{caron2021emerging}, it has been shown that self-supervised vision transformers \citep{dosovitskiy2020image} can perform zero-shot segmentation and that the embeddings of such models yield high KNN-classification accuracies. These have been improved by better training strategies and a large curated pretraining dataset in \citep{oquab2023dinov2}.
CLIP \citep{radford2021learning} demonstrated that the connection of images and text via an image and a text encoder, respectively, leads to high zero-shot accuracies. Moreover, ImageBind \citep{girdhar2023imagebind} demonstrated that robust joint embeddings can be achieved across different modalities without direct pairing, instead linking each modality to images.
\subsection{Prompting}
The systematic discovery of better prompts originated in the natural language modeling community; see, for example, \citep{jiang2020can, shin2020autoprompt, gao2020making, beyer2020we, liu2024gpt, lester2021power}.
Prompting has been successively applied to CLIP in CoOP \citep{zhou2022learning} and CoCoO \citep{zhou2022conditional}. Specifically, CoOp initializes the text prompts as a set of learnable embeddings, which is then continuously updated by forwarding it through the text encoder and applying gradient descent. This procedure is then refined in CoCoO to by learning prompts while fixing important words that highlight information about the class.
Discovering prompts for multi modal models remains an active area of research  \citep{zang2022unified, du2022learning, khattak2023maple, shen2024multitask}.
\subsection{Adapter Techniques}\label{adapter}
Adapter techniques, which freeze the original model weights and train only additional adapter layers, have gained popularity. Their main advantages are improved computational efficiency, since the backbone is not updated, and greater robustness to distribution shifts in few-shot settings. 
CLIP-Adapter \citep{gao2024clip} introduced residual MLP layers to incorporate new knowledge from few-shot data into the CLIP model. Subsequently, Tip-Adapter \citep{zhang2021tip} augmented this approach with a cache that stores feature maps from the few-shot visual encoder. 
In \citep{zhang2023prompt}, a cascade of foundation models was used to further enhance performance. Ta-Adapter \citep{zhang2024ta} fine-tuned the CLIP encoder to better encode task-specific knowledge. Meta-Adapter addressed the limitation of offline fine-tuning by proposing an adapter that learns online.
In \citep{gondal2024domain}, a method was proposed to align intermodal and intramodal embeddings. Proto-Adapter \citep{kato2024proto} extended Tip-Adapter by constructing class prototypes from the cached features.
In \citep{kato2024proto}, the adapter was simply constructed from class prototypes without any training.
\citet{bai2024advancing} already considered adapter ensembles in the context of image / text retrieval.
In \citep{zhang2023llama, gao2023llama} LLaMa-adapters where proposed for large language models such as the LLaMa herd \citep{touvron2023llama, touvron2023llama2, grattafiori2024llama}.

\subsection{fine-tuning and Robustness}\label{fine-tune}
Fine-tuning can greatly improve the performance of a model, but it can also result in less robustness to distribution shifts; see, for example, \citep{wortsman2022robust}, \citep{wortsman2022model}. We focus on distribution shifts where the class labels remain unchanged, but the visual appearance varies due to changes in conditions.
To address the robustness issue, several extensions of the ImageNet dataset were proposed in the literature. In \citep{recht2019imagenet}, four new ImageNet evaluation benchmarks were proposed. In \citep{hendrycks2021nae}, the ImageNet-A consisting of adversarial examples was proposed. The ImageNet-Sketch dataset was proposed in \citep{wang2019learning} and contains sketches of instances from ImageNet classes.
The ImageNet-R dataset was proposed in \citep{beyer2020we}.
In \citep{barbu2019objectnet}, ObjectNet, a bias-controlled dataset with several tasks has been proposed.
Although fine-tuning can result in less accuracy under distribution shifts, it has seen success in \citep{wortsman2022robust} where the original zero-shot weights of CLIP and the fine-tuned weights were linearly interpolated, resulting in a favorable in-distribution and out-of-distribution tradeoff.

Fine-tuning CLIP has seen success in \citep{wortsman2022robust} by interpolating between the original zero-shot weights and the fine-tuned weights. This effectively tackled the problem that a fine-tuned model usually performs poorly after a distribution shift.
Another improvement over plain fine-tuning was achieved in \citep{wortsman2022model} where the model was fine-tuned several times with different hyperparameters. The final model has been constructed by averaging the weights on all runs. This improved the model's robustness without adding additional inference cost.

In the natural language processing community, a dominant fine-tuning strategy LoRA \citep{hu2022lora} emerged. Instead of training all parameters of a model, only a lower rank subset of the entire parameter space is being trained during LoRA.
In \citep{ren2024melora}, an ensemble of low-rank adapters was used to fine-tune language models.
LoRA has also been applied to vision-language models such as CLIP in \citep{zanella2024low}. \citet{veasey2024parameter} applied LoRA to fine-tune DINOv2 on lung nodule classification.
\subsection{Domain Adaptation of DINOv2 models}
Compared to CLIP, significantly fewer studies have contributed to the domain adaptation of DINOv2. In \citep{deng2023universal} a baseline study about domain adaptation for DINOv2 and CLIP has been carried out. In \citep{englert2024exploring}, and \citep{kaplan2024domain} a study on adaptation to natural image segmentation tasks has been carried out. There exist several studies on the adaptation of DINOv2 to the medical domain. In \citep{huix2024natural} several natural image foundation models have been compared in terms of their adaptability to the medical domain. In \citep{zhao2024retrieval} and \citep{reddy2024data}, DINOv2 has been used for few-shot medical image segmentation. \citet{zhang2024learning} adapted DINOv2 to capsule endoscopy diagnosis using LoRA \citep{hu2021lora}. \citep{cui2024surgical} adapted the model to depth estimation in endoscopic surgery.
DINOv2 has also been adapted to other specialized domains such as visual place recognition \citep{lu2024towards}, change detection \citep{zhao2023adapting} and deep fake detection \citep{nguyen2024exploring}. \citep{englert2024exploring} and \citep{abedi2024euda} considered unsupervised domain adaptation. \citet{jose2024dinov2} applied locked-image tuning \citep{zhai2022lit} to DINOv2.
\subsection{Structural Reparameterization}\label{sec:rep}
Structural reparameterization is a technique in which models are trained using an architecture that is more optimization-friendly than the one used during inference. After training, the model is reparameterized to an equivalent architecture suitable for deployment.
This method gained popularity with {RepVGG}~\citep{ding2021repvgg}, where the classic VGG architecture~\citep{simonyan2014very} was augmented with skip connections during training. It had already been explored in {DiracNet}~\citep{zagoruyko2017diracnets} to train very deep convolutional networks without explicit skip connections. More recently, {FastViT}~\citep{vasu2023fastvit} used structural reparameterization to pretrain mobile-efficient vision transformers.

\section{Proposed Method}\label{sec:method}
Our method is based on the CLIP-Adapter~\citep{gao2024clip}. We will first review how classifiers can be generated to apply foundation models for classification in Section~\ref{sec:class}.
We then briefly review CLIP-Adapter in Section~\ref{sec:clip-adapter}. Finally, in Section~\ref{sec:cat}, we will introduce our method.
\subsection{Generation of Classifiers for Foundation Models}\label{sec:class}
To motivate our method, we will first recall how foundation models are usually adapted to classification tasks through KNN or prototypical classifiers.
We will consider foundation models that naturally form a pair of an image encoder and a prompt encoder. The image encoder (e.g. a CNN or a vision transformer \citep{dosovitskiy2020image}) projects an image $I \in \mathbb{R}^{3\times H \times W}$ onto a vector $x \in S^{D-1} = \lbrace x \in \mathbb{R}^D \ |\ |x| = 1 \rbrace$ with $H, W$ being the width and height of the image and $D$ the number of output features of the encoder.
Our method is designed to work independently of the prompt encoder's modality.
In the case of CLIP \citep{radford2021learning}, the prompt encoder is a language model, typically a transformer \citep{vaswani2017attention} which has been trained to produce embeddings of captions that align with the corresponding embeddings of the image encoder. In particular, natural language is used as the prompt modality in that case.
When using DINOv2 \citep{oquab2023dinov2}, the prompt encoder is the same model as the image encoder, so we may thus use natural images as prompts. This will in fact lead to a few-shot method, because training data is necessary to build the classifier, whereas CLIP prompts can simply be written down without any training data, leading to a zero-shot classifier.

There are two principal ways to construct a classifier using prompts.
The first one we shall describe is to form a prototype \citep{snell2017prototypical}, \citep{ji2020improved} of each class in the embedding space of the prompt encoder.
Suppose that $\mathbb{R}^N$ is a vector space containing the prompts. 
The prompt encoder maps a prompt $P \in \mathbb{R}^N$ to a vector $y \in S^{D-1}$. Let $C_1, \dots, C_m$ denote the classes and $P_{i, 1}, \dots, P_{i, r},$ the prompts corresponding to class $C_i$ (for example, texts like 'An image of a dog' if CLIP is being used or images of dogs if DINOv2 is being used).
The prototype corresponding to class $C_i$ is $p_i = \frac{1}{|\Tilde{p}_i|} \Tilde{p}_i \in S^{D-1}$ where $\Tilde{p}_i = \sum\limits_{j = 1}^r \ P_{i, j}$. Once the prototypes have been computed, the classifier can be constructed by setting $W = \begin{pmatrix}
    p_1,
    \dots,
    p_m
\end{pmatrix}^T \in \mathbb{R}^{m \times D}$. If $x \in S^{D-1}$ is the image embedding of a sample, the logits can be computed as \begin{equation}\label{prot}
    \mathrm{logits} = \frac{W \cdot x}{\tau} \in \mathbb{R}^m
\end{equation} and the highest logit indicates the class prediction.
The temperature $\tau > 0$ is a hyperparameter that controls the sharpness of the output distribution after applying the softmax function \[\mathrm{softmax}(z_1, \dots, z_m) = \frac{1}{\sum_{j = 1}^m e^{z_j}}\left( e^{z_1}, \dots, e^{z_m} \right).
\]
This technique has also been used in \citep{radford2021learning} to form zero-shot classifiers (zero-shot in the sense that a user can specify the classes in terms of language and use the model without extra training).

An alternative to using prototypes for classification, particularly for DINO-type models, is to use K-Nearest-Neighbors (KNN) to deploy the model.
If $P_{i_1, r_1}, \dots, P_{i_k, r_k}$ are the $k$ prompt encodings with the highest cosine similarity ($\mathrm{cosine-similarity}(x, y) = \langle x , y\rangle$) and $h_1, \dots, h_k$ are their one-hot encoded categories, i.e. $h_j = e_{i_j}$ (with $e_1,\dots, e_m$ denoting the standard basis vectors of $\mathbb{R}^m$), then \begin{align*}
    \mathrm{logits} = \sum\limits_{j = 1}^k \exp\left(\frac{1}{T} x \cdot P_{i_j, r_j}\right) h_j
\end{align*}
for a hyperparameter $T > 0$.
\subsection{Review of CLIP-Adapter}\label{sec:clip-adapter}
In this section, we will review the CLIP-Adapter \citep{gao2024clip} on which our method is based.
\begin{figure}[t]
  \centering
  \includegraphics[width=\linewidth]{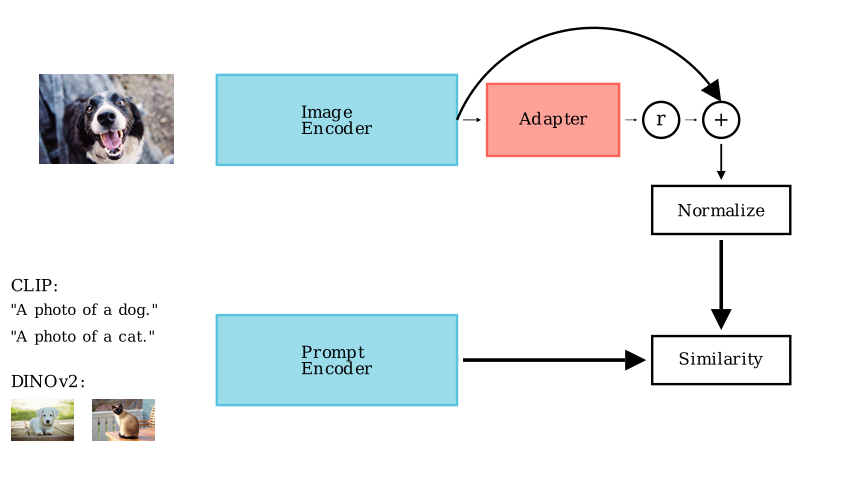}
  \caption{Overview of CLIP-Adapter. An additional adapter layer computes new features and adds these to the original features of the vision encoder before comparing with the text features. An extra hyperparameter $0 < r < 1$ called the residual ratio is used to scale the adapter features during inference, leading to an interpolation between the original and the adapted features.
  This method can be adopted to DINOv2 models by simply using training images as prompts.}
  \label{fig:clip-adapter}
\end{figure}
Figure~\ref{fig:clip-adapter} provides a visual overview of the CLIP adapter which we shall now describe mathematically in detail.
Suppose that $W_1 \in \mathbb{R}^{D/\mathrm{red} \times D}$, $W_2 \in \mathbb{R}^{D \times D/\mathrm{red}}$, $b_1 \in \mathbb{R}^{D/\mathrm{red}}, b_2 \in \mathbb{R}^D$ are a set of weights with $\mathrm{red}$ being a hyperparameter.
Then \begin{equation}\label{a}
    A_{W_1, W_2, b_1, b_2}(x) = 
    W_2 \cdot \sigma( W_1 \cdot x + b_1) + b_2
\end{equation}
where $x \in S^{D-1}$ are the features computed from the image encoder of the vision encoder. The function $\sigma$ is a nonlinear activation for which we use GeLU \citep{hendrycks2016gaussian}.
The final features are defined by
\begin{equation}\label{ratio}
    f = \frac{x + ar}{|x + ar|}
\end{equation}
with $a = A_{W_1, W_2, b_1, b_2}(x)$ according to \eqref{a} and $0 < r < 1$ a hyperparameter called the residual ratio, typically set to $1$ during training and tuned afterwards.
In order to train the adapter, one can use the original prototypes and formula~\eqref{prot} to form the logits. In this work, we will parameterize the temperature $\tau \in \mathbb{R}_+$ by $\tau = \exp(-\mathrm{scale})$ with $\mathrm{scale} \in \mathbb{R}$ being a hyperparameter which implies that $\mathrm{logits} = W \cdot x \ \exp(\mathrm{scale})$ with $W$ being the zero-shot weights described in Section~\ref{sec:class}.
The hyperparameter $r$ is very important. It allows one to interpolate between the original zero-shot/prototype prediction and the adapted prediction, which is important if the adapter overfits. In practice, sensitivity to the residual ratio is a bottleneck for many adapter methods because extensive evaluation splits may not be available in few-shot learning scenarios.

When adapting CLIP-Adapter to DINOv2, there is a design choice, because the prompts being used must be taken from the training dataset. 
One option is to simply compute the prompts once and accept that the current sample used for training is part of the prototype. The other is to compute the prototype at each training iteration and remove the current training sample from the corresponding prototype. We found that using the first strategy works better for very few shots such as $2, 4$ while the second one for higher numbers of shots such as $8, 16$. Thus, we adapt these strategies in the corresponding settings. We will reference to these as 'no-mask' and 'mask', respectively.
For simplicity, from now on we will use the shorthand adapter for CLIP-Adapter which can either be used for CLIP or DINOv2.
\subsection{Soup-Adapters}\label{sec:cat}
Our Soup-Adapter method is essentially the adapter equivalent to the uniform averaging strategy in model soups \citep{wortsman2022model}.
More specifically, we train several adapters called the components with weights $W_1^j, W_2^j, b_1^j, b_2^j$ for $j \in \lbrace 1, \dots, K \rbrace$ and set \begin{equation}
    a = \frac{1}{K} \sum\limits_{j=1}^K A_{W_1^j, W_2^j, b_1^j, b_2^j}(x)
\end{equation}
and compute the final features again according to \eqref{ratio}.
We vary different hyperparameters such as the $\mathrm{red}$-factor and the learning rate when training individual adapters.

\begin{figure}[t]
  \centering
  \includegraphics[width=\linewidth]{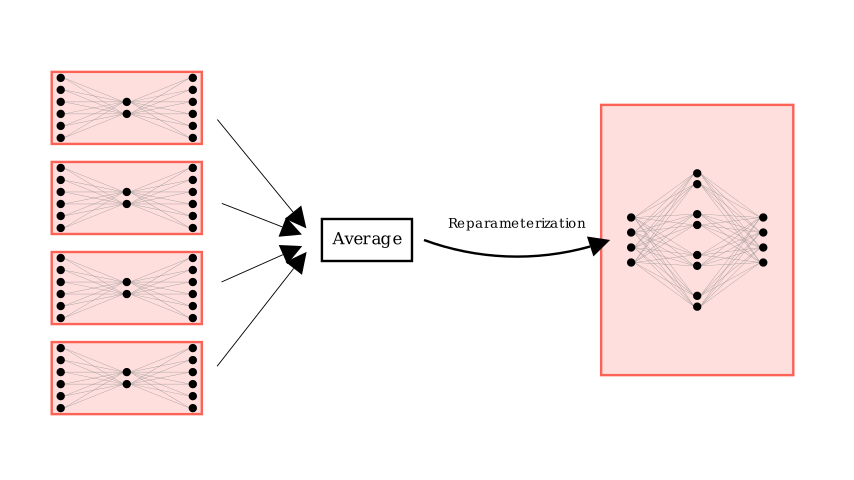}
  \caption{Reparameterization of an ensemble of adapters to a single one using concatenation. The resulting dimension of the hidden layer adds up.}
  \label{fig:rep}
\end{figure}

In practice, the parameterization above could lead to slower inference times because of the need to compute $K$ individual adapter outputs. However, this issue can be resolved by reparameterization which is visually depicted in Figure~\ref{fig:rep}.
To be more precise,
\begin{equation}\label{cat}
    \begin{split}
        W_1 = \begin{pmatrix}
        W_1^1 \\
        \vdots \\
        W_1^K
    \end{pmatrix}, W_2 = \frac{1}{K}\begin{pmatrix}
        W_2^1 &\dots & W_2^K
    \end{pmatrix} \\
    b_1 = \begin{pmatrix}
        b_1^1\\
        \vdots \\
        b_1^K
    \end{pmatrix}, b_2 = \frac{1}{K}\begin{pmatrix}
        b_2^1 &\dots & b_2^K
    \end{pmatrix}
 HG   \end{split}
\end{equation}
will yield an adapter with $A_{W^1, W^2, b^1, b^2}(x) = a(x)$ for any $x \in S^{D-1}$.
Due to the division by the $red \in \mathbb{N}$-factors, this new set of parameters will typically still be relatively small compared to the number of parameters of the model itself.
Model soups~\citep{wortsman2022model} also considered a greedy strategy of averaging the weights of independently fine-tuned models. The greedy averaging strategy first sorts the components via descending accuracy on the validation set. The soup is then constructed by adding the components to the soup sequentially and only keeping them if the accuracy increases. However, this strategy is not feasible for few-shot learning scenarios, because large validation datasets are typically not available.
We will use just one residual ratio hyperparameter for the Soup-Adapter and scale each component by this residual ratio.
\section{Experiments}\label{sec:experiments}
\subsection{Datasets}\label{sec:data}
We evaluated our method on multiple datasets commonly encountered in few-shot learning.
Specifically, we use ImageNet~\citep{deng2009imagenet},  StandfordCars~\citep{krause20133d},  UCF101~\citep{soomro2012dataset},  Caltech101~\citep{fei2004learning}, Flowers102~\citep{nilsback2008automated}, SUN397~\citep{xiao2010sun},  DTD~\citep{cimpoi2014describing}, 
EuroSAT~\citep{helber2019eurosat}, 
FGVCAircraft~\citep{maji2013fine}, 
OxfordPets~\citep{parkhi2012cats}, 
FOOD101~\citep{anderson2018bottom}.
To evaluate the robustness of our method, we also consider the average accuracy of four ImageNet evaluation sets.
These are ImageNetV2~\citep{recht2019imagenet}, ImageNet-A~\citep{hendrycks2021many}, ImageNet-R~\citep{beyer2020we} and ImageNet-Sketch~\citep{wang2019learning}.
\subsection{Implementation Details}
We carry out all experiments using the PyTorch deep learning framework. We use the ViT \citep{dosovitskiy2020image} variants of CLIP. Most of the experiments are carried out using the ViT-B/32 CLIP model and the DINOv2 ViT-B/14 variant.
We use $N_{shot} = 2, 4, 8, 16$ to evaluate our method with different numbers of shots. For each individual adapter, we assign a new random seed and a randomly chosen hyperparameter configuration. The $\mathrm{red}$-factor is a random natural number between $2$ and $10$. The learning rate is chosen randomly from $2e-3, 1e-3, 5e-4$. As data augmentation, we apply a random resized crop with minimum scale $s \cdot 0.2 + (1 - s) \cdot 0.8$, random horizontal flips with probability $0.5$, color  jitter with brightness $0.4$, contrast $0.4$, saturation $0.2$, hue $0.1$ with probability $s \cdot 0.5$ and random grayscale with probability $s \cdot 0.2$ where $s$ is chosen randomly from $0.25, 0.5, 0.75, 1$.
Finally, the weight decay is randomly chosen from $1e-3, 1e-2, 5e-2$.
We use an input size of $224\times224$ for CLIP for both evaluation and training, an input size of $308\times 308$ for DINOv2 during evaluation and $224\times 224$ during training.
The Soup-Adapter is evaluated with a single residual ratio $0 < r <1$ for the combined adapter as described by formula~\eqref{cat}. The network is trained using cross-entropy with label smoothing~\citep{hu2018squeeze} $0.1$.
We typically use a batch size of $32$. The number of necessary epochs strongly depends on the dataset and whether CLIP or DINOv2 is used and can vary between $8$ and $300$. All experiments were carried out on an NVIDIA-GeForce RTX 3090 GPU, a random seed and deterministic algorithms in PyTorch were used to guarantee reproducibility. Unless otherwise stated, we always consider $K = 8$ (i.e. the Soup-Adapter is formed out of $8$ individual components).
\footnote{Our code is available for reproduction: https://github.com/trawler0/Soup-Adapter/}
\subsection{Baselines}
For CLIP, we consider the zero-shot CLIP variant as described in Section~\ref{sec:class}. The prompts used are the original ones used by CLIP~\citep{radford2021learning} augmented by additional ones.
For DINOv2, we used the embeddings of the available labeled samples to form the prototype of each class. To ensure consistency, we use the same prototypes for adapters and consequently also for Soup-Adapters as used in the baselines. This also implies that our method applied with the residual ratio $r=0$ at inference corresponds to the prototype baseline.
For DINOv2, we additionally consider the KNN evaluation with $k=10$ and $\tau = 0.1$, which is a common evaluation strategy used in the DINO~\citep{caron2021emerging} based work. However, we point out that the prototypical evaluation typically performed better, particularly in terms of robustness in our experiments. Finally, we consider the average accuracy of the individual adapters used to form the Soup-Adapter as a baseline.
\subsection{Results}
\begin{figure*}
    \centering
    \includegraphics[width=1\textwidth]{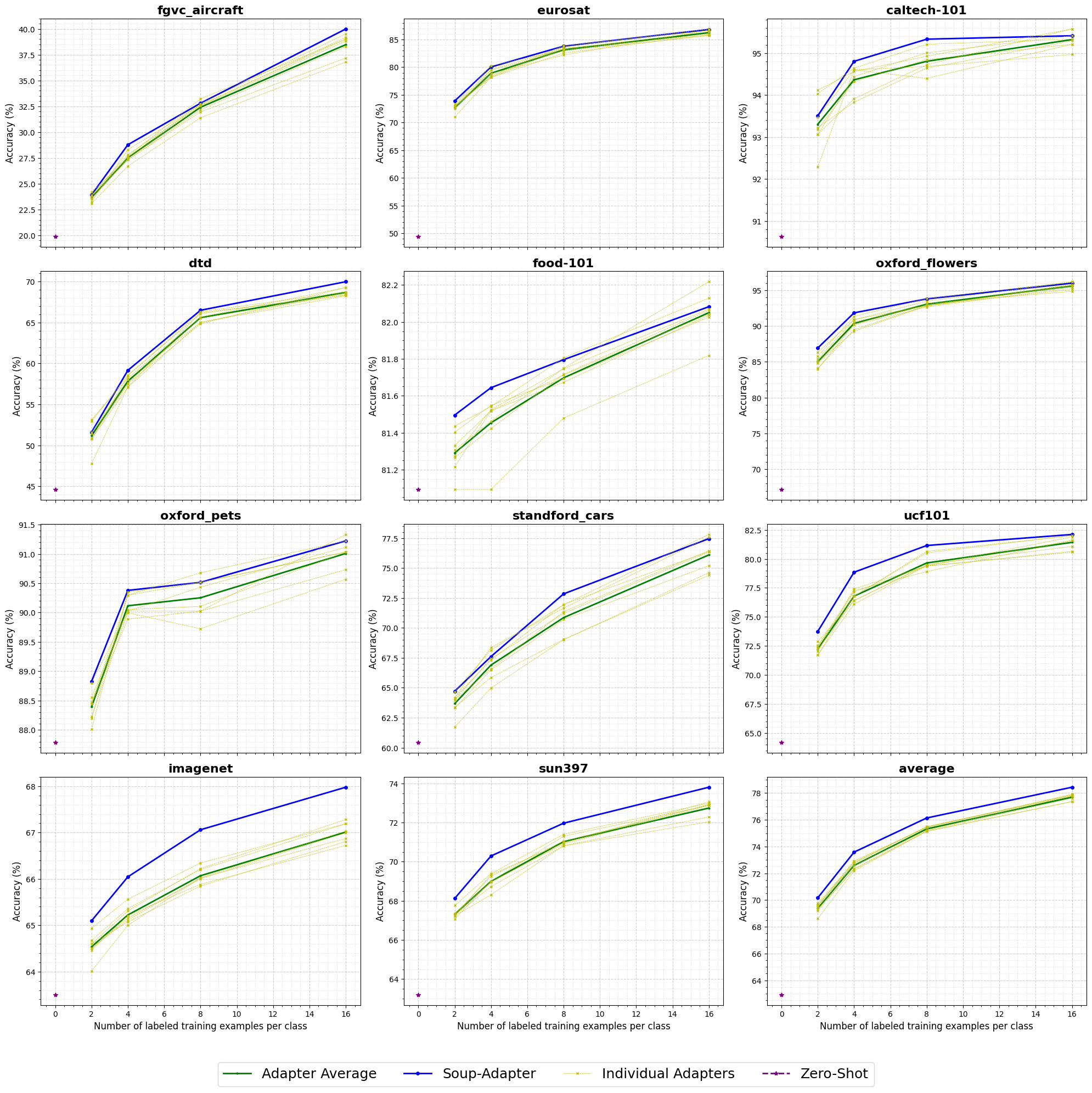}
    \caption{The effect of Soup-Adapter on in-distribution accuracy for CLIP models. Individual adapters with varying hyperparameters are shown in yellow. The green line represents the average accuracy of the $10$ individual adapters. The Soup-Adapter is shown in blue. The zero-shot baseline is represented by the purple star.}
    \label{fig:clip}
\end{figure*}

\begin{figure*}
    \centering
    \includegraphics[width=1\textwidth]{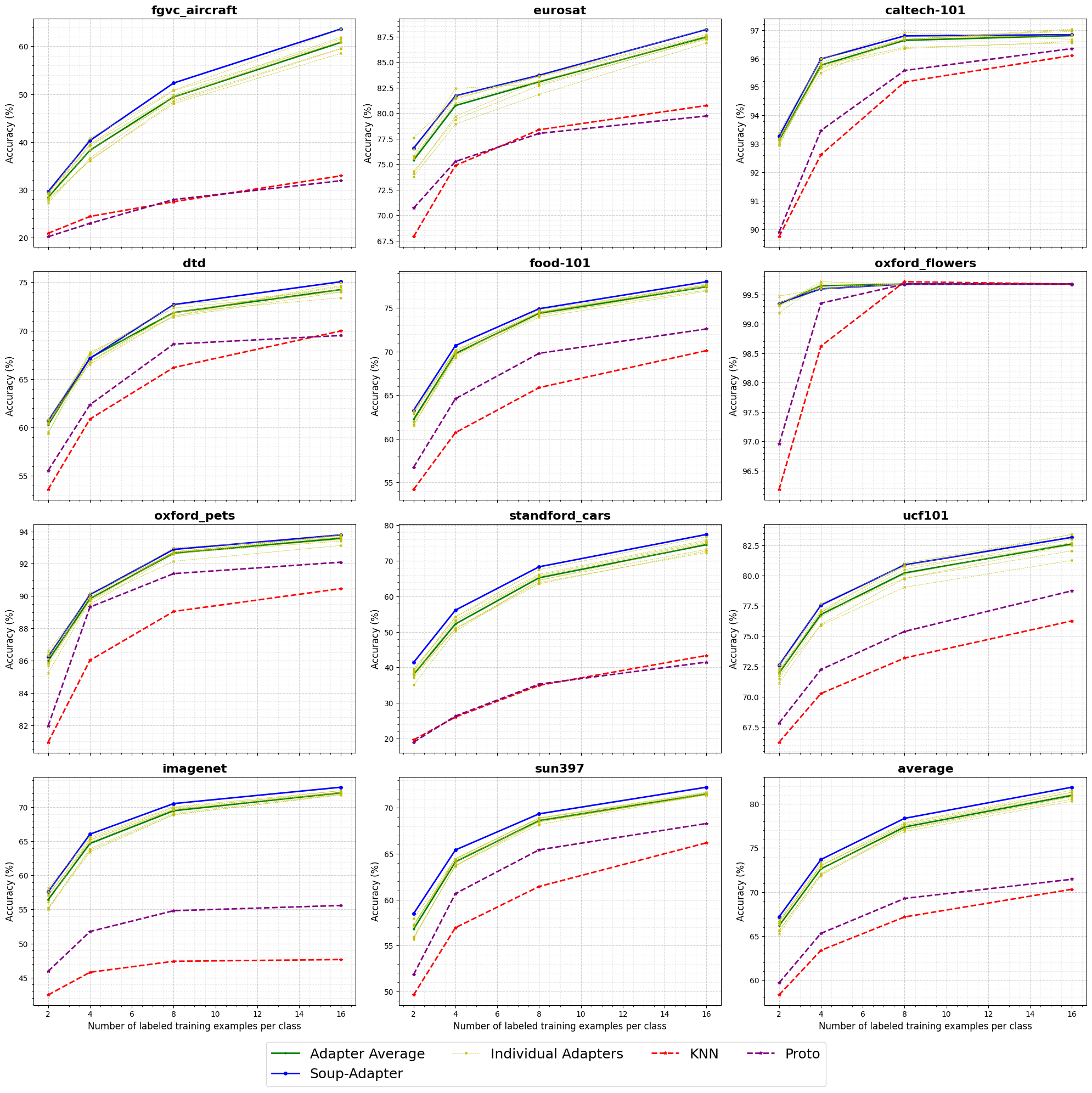}
    \caption{The effect of Soup-Adapter on in-distribution accuracy for DINOv2 models. Individual adapter with varying hyperparameters are shown in yellow. The green line represents the average accuracy of these $10$ individual CLIP-Adapters. The Soup-Adapter is shown in blue. The proto and KNN baseline are represented by the purple and red lines, respectively.}
    \label{fig:dino}
\end{figure*}

We will now discuss the main results of this work. Starting with CLIP, Figure~\ref{fig:clip} shows that Soup-Adapter significantly outperforms the individual components. Moreover, even in setups where there is a high variance across individual runs, the Soup-Adapter typically achieves performance comparable with the best individual runs. This is a very desirable property in few-shot learning, where extensive evaluation datasets are often not available, making hyperparameter tuning difficult in practice. However, we see that the performance gain due to Soup-Adapter tends to decrease if there is less variance across individual adapter runs. 
It is evident that the use of Soup-Adapter effectively overcomes the high variance between individual runs highlighted in yellow.
Figure~\ref{fig:dino} shows the results obtained for the DINOv2 models. We see that the prototypical baseline typically outperforms the KNN evaluation, which is interesting, because DINO \citep{jose2024dinov2, caron2021emerging} models are typically evaluated with KNN instead of prototypes. Furthermore, adapters outperform these baselines by a large margin.
We see a smaller gap between the Soup-Adapter and individual runs than for CLIP models.

\begin{figure*}
    \centering
    \includegraphics[width=1\linewidth]{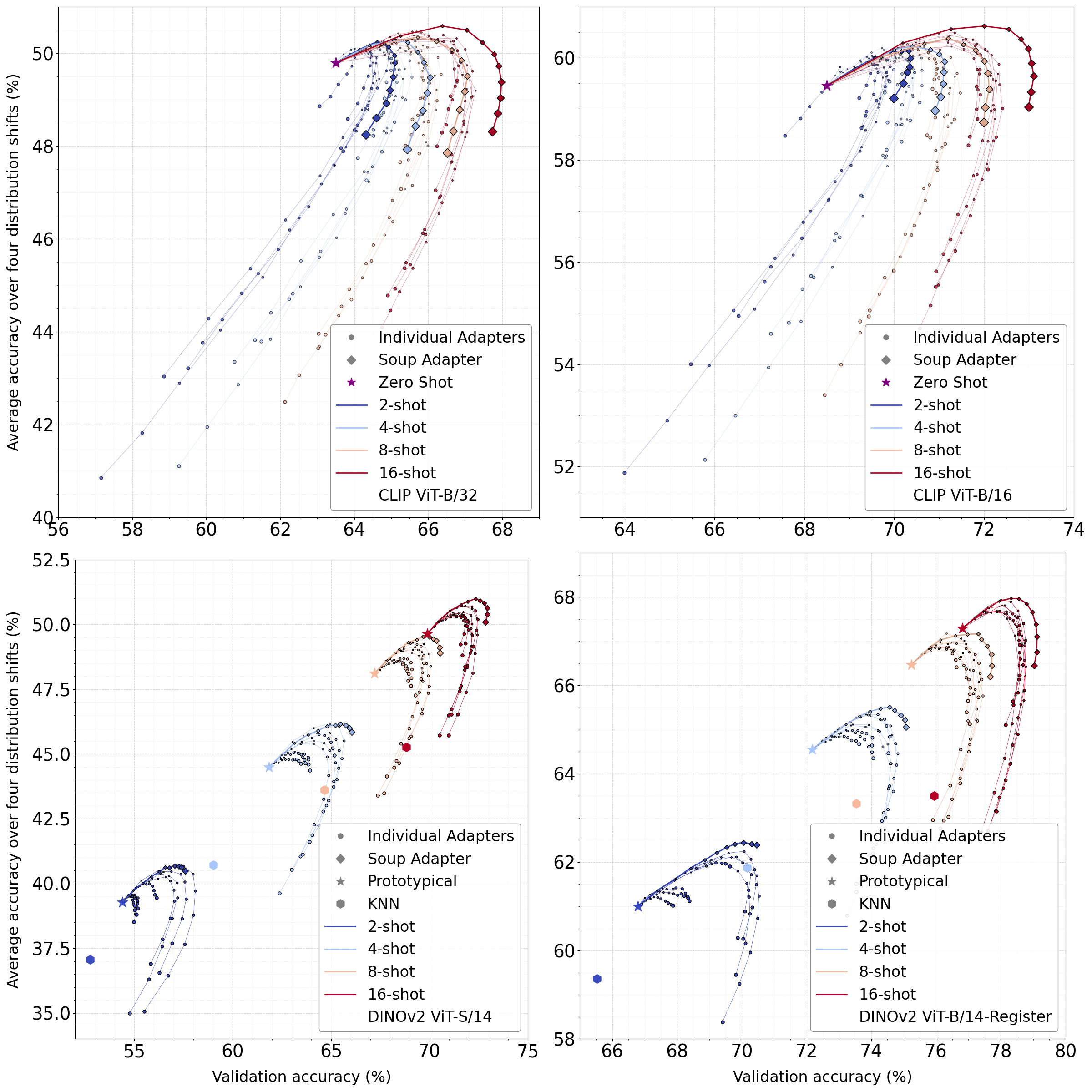}
    \caption{Analysis of Soup-Adapter's robustness under distribution shifts on the ImageNet dataset with CLIP models on top and DINOv2 on the bottom.
    The x-axis shows the in-distribution accuracy and the y-axis the out-of-distribution accuracy. Individual curves are obtained by varying the residual ratio $r$ between $0$ and $1$ in $0.1$-steps.}
    \label{fig:robust}
\end{figure*}

This might lead to the wrong impression that the Soup-Adapter does not yield any benefit for DINOv2 models. However, this is not true because Soup-Adapters increase the robustness under distribution shift, which can be seen in Figure~\ref{fig:robust}. 
For both CLIP and DINOv2, we vary the residual ratio $0 < r< 1$, then plot the in-distribution accuracy on the x-axis and out-of-distribution accuracy on the y-axis. For CLIP-models, we see that a too high residual ratio can be detrimental, often leading to worse than zero-shot results, particularly under distribution shifts. Although this phenomenon is not as severe for DINOv2, we observe that the Soup-Adapter robustness curve is also favorable for those models. The optimal in-distribution score of the Soup-Adapter is often very close to the score of individual adapters, which is consistent with the earlier findings when we inspected the Figure~\ref{fig:dino}. But the accuracy under a distribution shift can vary a lot, and Soup-Adapter provides a favorable curve. These plots also verify that Soup-Adapters are much less sensitive to the residual ratio when distribution robustness is considered, although it is still an important hyperparameter. We can also see that due to the different approach to create prototypes for DINOv2 compared to CLIP, the graphs look fundamentally different. With more shots, the prototypical baseline smoothly increases the model's accuracy, whereas the zero-shot baseline for CLIP achieves a fixed score. The domain adaptation of CLIP typically has the potential to yield a larger performance boost, but it is also subject to more variance with improper hyperparameter tuning and a poorly chosen residual ratio. We can clearly see that the Soup-Adapter is extremely helpful in solving these problems.
\subsection{Ablations}
In this section, we will present several ablations on our method. An important question is how large $K$ (number of individual adapters in the Soup-Adapter) should be.  
\begin{figure}[!]
    \centering
    \includegraphics[width=1\linewidth]{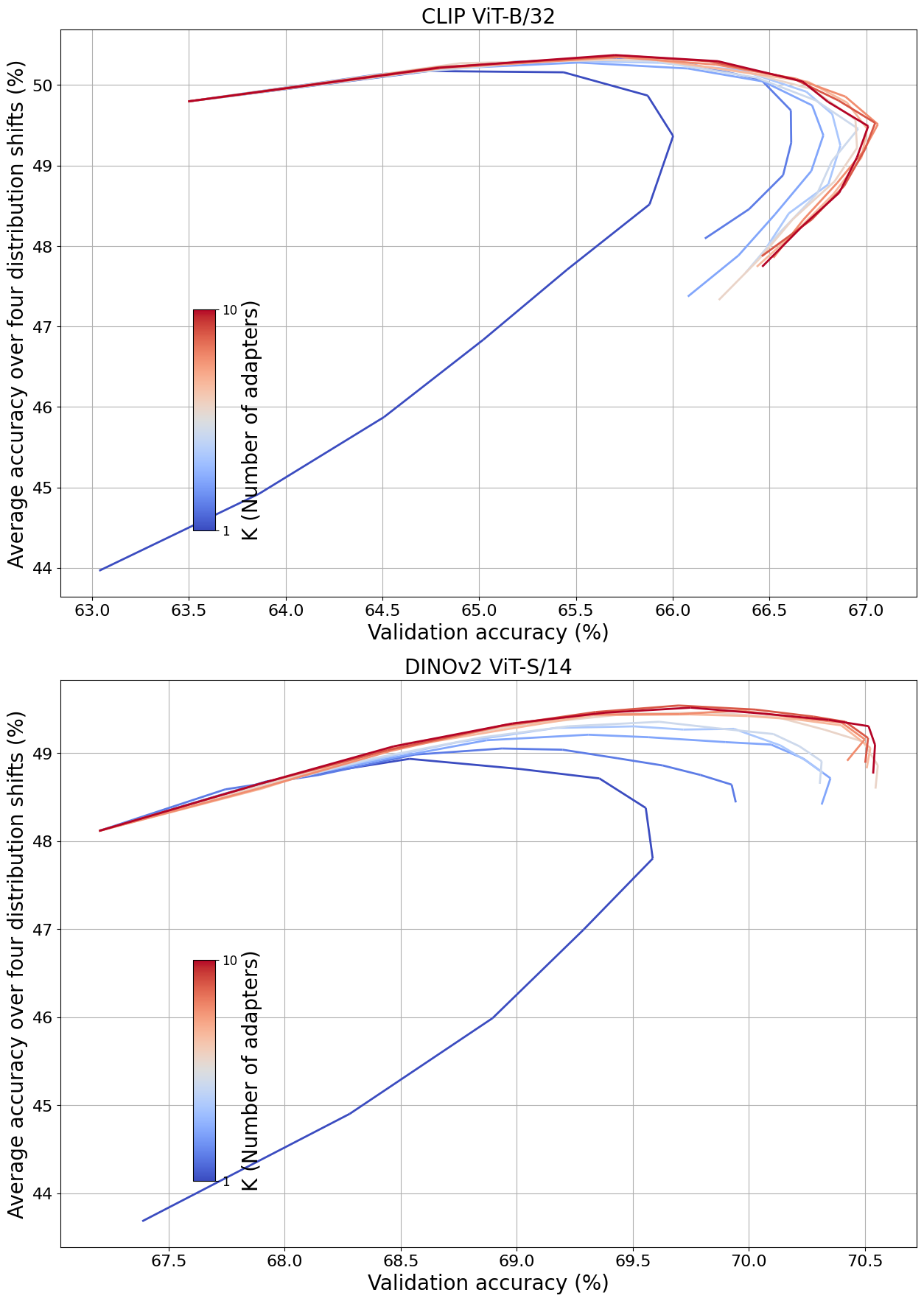}
    \caption{Ablation on the number of number of indiviudal adapters in the Soup-Adapter using the ImageNet $8$-shot evaluation.}
    \label{fig:ablate}
\end{figure}
We can clearly see in Figure~\ref{fig:ablate} that even $K=3$ or $K=4$ already significantly increases the robustness and decreases the dependency on the residual ratio. We can also see that $K=10$ sufficed in this experiment to reach saturation with respect to this performance boost.

Table~\ref{table:mask} contains an ablation of the choice we made to mask the instances currently seen during the training of the prototype for DINOv2.
\begin{table*}[t]
\caption{Accuracy (\%) on FGVCAircraft for different residual ratios and configurations for DINOv2 with $N=8$ shots. Soup-Adapters clearly outperform the average over $8$ individual adapter runs in almost all setting. We can see that the mask strategy (mask the current training instance in the prototype) performs better if very few shots such as $2, 4$ are available. On the other hand, the mask strategy outperforms the no-mask strategy for the higher numbers of shots $8, 16$.}
\label{table:mask}
\centering
\footnotesize
\setlength{\tabcolsep}{5pt}
\begin{tabular}{lcccccccc}
\toprule
& \multicolumn{2}{c}{\textbf{2-shot}} & \multicolumn{2}{c}{\textbf{4-shot}} & \multicolumn{2}{c}{\textbf{8-shot}} & \multicolumn{2}{c}{\textbf{16-shot}} \\
\cmidrule(lr){2-3} \cmidrule(lr){4-5} \cmidrule(lr){6-7} \cmidrule(lr){8-9}
& \textbf{mask} & \textbf{no mask} & \textbf{mask} & \textbf{no mask} & \textbf{mask} & \textbf{no mask} & \textbf{mask} & \textbf{no mask} \\
\midrule
avg. ind. adapter $r=0.3$ & 22.5 & 22.9 & 29.0 & 28.0 & 35.7 & 34.0 & 42.9 & 41.3 \\
avg. ind. adapter $r=0.5$ & 23.6 & 23.9 & 33.0 & 30.9 & 41.0 & 38.5 & 50.5 & 48.0 \\
avg. ind. adapter $r=0.7$ & 24.8 & 26.3 & 36.6 & 34.3 & 47.2 & 44.3 & 58.2 & 55.3 \\
avg. ind. adapter $r=1.0$ & 25.4 & 29.6 & 39.3 & 39.1 & 52.6 & 52.5 & 65.4 & 64.3 \\
Soup-Adapter $r=0.3$ & 22.7 & 23.2 & 28.9 & 27.8 & 35.8 & 34.1 & 43.0 & 41.3 \\
Soup-Adapter $r=0.5$ & 23.6 & 24.0 & 33.0 & 30.7 & 41.2 & 38.4 & 54.6 & 48.4 \\
Soup-Adapter $r=0.7$ & 24.5 & 25.9 & 37.2 & 34.4 & 47.8 & 44.6 & 58.3 & 55.6 \\
Soup-Adapter $r=1.0$ & 25.6 & 30.0 & 41.1 & 39.9 & 54.6 & 54.3 & 67.3 & 66.1 \\
\bottomrule
\end{tabular}
\end{table*}
We can see that this choice is favorable for the $8$ and $16$-shot, but not for the $2$ and $4$-shot.

\begin{table}[t]
\caption{Performance comparison between the prototypical baseline for DINOv2, the average of $4$ adapters and a Soup-adapter with $K=4$ components, both with residual ratio $r=1$ in the $4$-shot setting. This demonstrates that adapters yield performance improvements for very large architectures such as ViT-G/14 with $1.1$ billion parameters. We can also see that Soup-Adapters help to see even better performance gains.}
\label{tab:dinov2-g}
\centering
\footnotesize
\setlength{\tabcolsep}{3pt}
\begin{tabular}{lcccc}
\toprule
 & G14 & L14 & B14 & S14 \\
\midrule
\textbf{val, Prototypical} $(\%)$  & 73.69 & 73.78 & 72.18 & 61.84 \\
\textbf{val, avg. ind. Adapter} $(\%)$  & 75.76 & 75.53 & 73.82 & 63.46 \\
\textbf{val, Soup-Adapter} $(\%)$   & 76.60 & 76.75 & 74.94 & 65.91 \\
\midrule
\textbf{v2, Prototypical} $(\%)$   & 67.93 & 67.19 & 65.48 & 52.66 \\
\textbf{v2, avg. ind. Adapter} $(\%)$   & 68.96 & 68.34 & 66.07 & 54.09 \\
\textbf{v2, Soup-Adapter}  $(\%)$   & 69.46 & 69.54 & 67.23 & 56.12 \\
\midrule
\textbf{sketch, Prototypical} $(\%)$ & 60.30 & 58.09 & 53.37 & 35.00 \\
\textbf{sketch, avg. ind. Adapter} $(\%)$ & 59.81 & 57.41 & 52.49 & 32.69 \\
\textbf{sketch, Soup-Adapter} $(\%)$ & 60.78 & 58.61 & 53.99 & 35.24 \\
\midrule
\textbf{IN-A, Prototypical} $(\%)$ & 78.21 & 75.05 & 68.03 & 38.56 \\
\textbf{IN-A, avg. ind. Adapter} $(\%)$ & 76.05 & 73.31 & 64.84 & 36.47 \\
\textbf{IN-A, Soup-Adapter} $(\%)$  & 77.21 & 74.92 & 67.12 & 39.56 \\
\midrule
\textbf{IN-R, Prototypical} $(\%)$ & 80.93 & 77.95 & 71.34 & 51.76 \\
\textbf{IN-R, avg. ind. Adapter} $(\%)$ & 77.73 & 75.43 & 67.94 & 47.37 \\
\textbf{IN-R, Soup-Adapter} $(\%)$  & 79.34 & 77.19 & 69.99 & 50.72 \\
\bottomrule
\end{tabular}
\end{table}
\begin{table}[t]
\caption{Performance comparison between the zero-shot baseline for CLIP, the average of $6$ adapters and a Soup-adapter with $K=6$ components, both with residual ratio $r=0.6$ in the $4$-shot setting. This demonstrates that adapters yield performance improvements for larger architectures such as ViT-L with $304$ million parameters. We can also see that Soup-Adapters help to see even better performance gains.}
\label{tab:clip-l}
\centering
\footnotesize
\setlength{\tabcolsep}{3pt}
\begin{tabular}{lcccc}
\toprule
 & L14@336 & L14 & B16 & B32 \\
\midrule
\textbf{val, Prototypical} $(\%)$  & 76.53 & 75.57 & 68.50 & 63.50 \\
\textbf{val, avg. ind. Adapter} $(\%)$  & 77.49 & 76.56 & 69.92 & 64.66 \\
\textbf{val, Soup-Adapter} $(\%)$   & 78.65 & 77.65 & 70.92 & 65.89 \\
\midrule
\textbf{v2, Prototypical} $(\%)$   & 70.92 & 69.86 & 61.91 & 55.96 \\
\textbf{v2, avg. ind. Adapter} $(\%)$   & 71.25 & 69.93 & 63.03 & 56.53 \\
\textbf{v2, Soup-Adapter}  $(\%)$   & 72.40 & 71.17 & 63.97 & 57.58 \\
\midrule
\textbf{sketch, Prototypical} $(\%)$ & 60.98 & 59.59 & 48.26 & 42.33 \\
\textbf{sketch, avg. ind. Adapter} $(\%)$ & 59.46 & 58.46 & 47.76 & 41.19 \\
\textbf{sketch, Soup-Adapter} $(\%)$  & 60.40 & 59.46 & 48.72 & 42.24 \\
\midrule
\textbf{IN-A, Prototypical} $(\%)$ & 77.48 & 70.77 & 49.95 & 31.55 \\
\textbf{IN-A, avg. ind. Adapter} $(\%)$  & 75.32 & 68.43 & 48.27 & 29.71 \\
\textbf{IN-A, Soup-Adapter} $(\%)$  & 76.49 & 69.53 & 49.01 & 30.32 \\
\midrule
\textbf{IN-R, Prototypical} $(\%)$ & 89.03 & 87.84 & 77.71 & 69.35 \\
\textbf{IN-R, avg. ind. Adapter} $(\%)$ & 87.49 & 86.13 & 75.85 & 66.97 \\
\textbf{IN-R, Soup-Adapter} $(\%)$  & 88.50 & 87.09 & 76.86 & 68.08 \\
\bottomrule
\end{tabular}
\end{table}
DINOv2 models are known for their universal features. Indeed, \citet{oquab2023dinov2} show that with enough data, linear evaluation and KNN-evaluation can be competitive with full fine-tuning for large DINOv2 encoders. 

In Table~\ref{tab:dinov2-g}, we can see that even with these large architectures adapters and particularly Soup-Adapters can help to significantly enhance the performance of larger DINOv2 models in few-shot adaptation setups. 

A similar trend can be observed for larger CLIP models such as ViT-L with $304$ million parameters, as shown in Table~\ref{tab:clip-l}. In this case, the Soup-Adapter yields an even higher relative benefit compared to individual adapters.

We have already seen in Figure~\ref{fig:robust} that the residual ratio is a crucial hyperparameter if the target domain is shifted. In Table~\ref{tab:best-ratio}, we can also see that on the set of evaluated data sets considered in $11$, the Soup-Adapter depends less on the optimal residual ratio than its components. Indeed, we can see that lower residual ratios typically tend to be better when very few shots such as $2, 4$ are available. With more available shots, it becomes more likely that a higher residual ratio performs better. This trend is true, both for the average score of the individual adapters and for the score of the Soup-Adapter. When individual adapters are evaluated with $2$ shots the performance drop with residual ratio $r=0.9$ instead of a tuned residual ratio is $3.6 \%$ compared to only $2.2\%$ when using Soup-Adapter. The difference decreases with more shots, but for $8$ shots using individual adapters, the difference between the score with a tuned residual ratio and $r=0.9$ is $1.6 \%$ compared to $1.0 \%$ when Soup-Adapter is used. This shows that Soup-Adapter is also less dependent on the residual ratio when evaluated in-distribution.

\begin{table}[t]
\caption{Average accuracy over $11$ datasets for CLIP models at various residual ratios. The best ratios are selected using the validation dataset. The average is computed after tuning the residual ratio on the validation dataset for each individual adapter.}
\label{tab:best-ratio}
\centering
\footnotesize
\begin{tabular}{lcccc}
\toprule
Method & 2 & 4 & 8 & 16 \\
\midrule
avg single adapter, best $r$ $(\%)$        & 69.4 & 72.6 & 75.3 & 77.7 \\
avg single adapter, $r = 0.3$ $(\%)$       & 68.5 & 71.5 & 73.7 & 75.8 \\
avg single adapter, $r = 0.6$ $(\%)$          & 67.6 & 71.6 & 74.5 & 77.0 \\
avg single adapter, $r = 0.9$ $(\%)$          & 65.9 & 70.3 & 73.7 & 76.5 \\
Soup-Adapter, best $r$ $(\%)$           & 70.2 & 73.6 & 76.1 & 78.4 \\
Soup-Adapter, $r = 0.3$ $(\%)$          & 69.4 & 72.5 & 74.9 & 77.1 \\
Soup-Adapter, $r = 0.6$ $(\%)$          & 69.3 & 73.1 & 75.7 & 78.0 \\
Soup-Adapter, $r = 0.9$ $(\%)$          & 68.2 & 72.1 & 75.1 & 77.6 \\
\bottomrule
\end{tabular}
\end{table}
\section{Conclusion}\label{sec:conclusion}
In this work, we introduced Soup-Adapter, an ensemble of adapters. We have demonstrated that Soup-Adapters can enhance the performance of adapters. Additionally, Soup-Adapter addresses key limitations of few-shot learning methods, it makes hyperparameter tuning easier and increases the robustness under distribution shifts. We demonstrated that CLIP-Adapter-based methods, especially the proposed Soup-Adapter, perform effectively with DINOv2 models. The proposed strategy is generic and might be helpful whenever adapters are applied to foundation models. An interesting direction for future work is the use of Soup-Adapters for language models using LLaMa-adapters \citep{zhang2023llama, gao2023llama}.

To the best of our knowledge, this study, being an improvement of the well-established CLIP-Adapter method, does not introduce significant new risks to society. Potential risks such as biases present in the underlying data have already been recognized and discussed extensively in existing literature.

\textbf{Data Availability Statement:} No new data has been created to conduct this study. We used the publicly available data from the literature ~\citep{deng2009imagenet,krause20133d,  soomro2012dataset, fei2004learning, nilsback2008automated, xiao2010sun,  cimpoi2014describing, 
helber2019eurosat, 
maji2013fine, 
parkhi2012cats, 
anderson2018bottom, recht2019imagenet, hendrycks2021many, beyer2020we, wang2019learning}.

\bibliography{sn-bibliography}

\end{document}